\documentclass[a4paper,11pt]{amsart}

\usepackage{amstext}
\usepackage{amsthm}
\usepackage{amssymb}
\usepackage{color}
\usepackage{esint}
\usepackage{dsfont}
\usepackage{upgreek}
\usepackage{algorithm}
\usepackage{amsthm}
\usepackage{enumerate}
\usepackage{amsfonts}
\usepackage{amssymb}
\usepackage{mathtools}
\usepackage{braket}
\usepackage{bm}
\usepackage{amsmath}
\usepackage{graphicx}
\usepackage{caption}
\usepackage{subcaption}
\usepackage{fullpage}
\usepackage{longtable}
\usepackage{mathrsfs}
\usepackage{mathtools}
\usepackage{xcolor}
\usepackage{tikz}
\usepackage{array}
\usepackage{tabularx}

\numberwithin{equation}{section}
\graphicspath{ {./images/} }

\providecommand{\U}[1]{\protect\rule{.1in}{.1in}}
\newtheorem{theorem}{Theorem}[section]

\newtheorem{remark}[theorem]{Remark}

\DeclareMathOperator*{\minn}{min}

\definecolor{mygreen}{rgb}{0.1,0.75,0.2}

\newcommand{\nc}{\normalcolor}

\newcommand{\EE}{\mathcal{E}}

\newcommand{\R}{\mathbb{R}}
\newcommand{\G}{\mathcal{G}}

\title{ Traditional and accelerated gradient descent for neural architecture search }

\author{Nicol\'as Garc\'ia Trillos, F\'elix Morales, and  Javier Morales}
\address{Nicol\'as Garc\'ia Trillos, Department of Statistics, University of Wisconsin-Madison. 1300 University Avenue, Madison, WI, USA 53706  \\}
\email{garciatrillo@wisc.edu}

\address{ F\'elix Morales, Departmento de Ingener\'ia Inform\'atica, Universidad Cat\'olica Andr\'es Bello, Caracas 1000, Capital District, Venezuela \\}
\email{famorales.14@est.ucab.edu.ve}

\address{Javier Morales, Center for Scientific Computation and Mathematical Modeling (CSCAMM), University of Maryland, College Park MD 20742  \\}
\email{javierm1@cscamm.umd.edu}

\begin{document}

	\thanks{{\bf Acknowledgements:} N. Garc\'ia Trillos was supported by NSF-DMS 2005797. The work of J. Morales was supported by NSF grants DMS16-13911, RNMS11-07444 (KI-Net) and ONR grant N00014-1812465. The authors are grateful to the CSCAMM, to Prof. P.-E. Jabin and Prof. E. Tadmor for sharing with them the computational resources used to complete this work. Support for this research was provided by the Office of the Vice Chancellor for Research and Graduate Education at the University of Wisconsin-Madison with funding from the Wisconsin Alumni Research Foundation. }

	\maketitle

\begin{abstract}
	In this paper we introduce two algorithms for neural architecture search (NASGD and NASAGD) following the theoretical work by two of the authors \cite{N1} which used the geometric structure of optimal transport to introduce the conceptual basis for new notions of traditional and accelerated gradient descent algorithms for the optimization of a function on a semi-discrete space. Our algorithms, which use the network morphism framework introduced in \cite{Hillclimbing} as a baseline, can analyze forty times as many architectures as the hill climbing methods \cite{Hillclimbing,Graphsearch} while using the same computational resources and time and achieving comparable levels of accuracy. For example, using NASGD on CIFAR-10, our method designs and trains networks with an error rate of 4.06 in only 12 hours on a single GPU. 
	
\end{abstract}


\section{Introduction}
Motivated by the success of neural networks in applications such as image recognition and language processing, in recent years practitioners and researchers have devoted great efforts in developing computational methodologies for the automatic design of neural architectures, in order to use deep learning methods in further applications. Roughly speaking, most approaches for \textit{neural architecture search} (NAS) found in the literature build on ideas from reinforcement learning, evolutionary algorithms, and hill-climbing strategies-- see section \ref{sec:literature} for a brief review of the literature. All of these approaches attempt to address a central difficulty: the high computational burden of training multiple architecture models. Several developments in the design of algorithms, implementation and computational power, have resulted in methodologies that are able to produce neural networks that outperform the best networks designed by humans. Despite all the recent exciting computational developments in NAS, we believe that it is largely of interest to propose sound mathematical frameworks for the design of new computational strategies that can better explore the architecture space and ultimately achieve higher accuracy rates in learning while reducing computational costs. 

In this paper we propose two new algorithms for NAS: \textit{neural architecture search gradient descent} \textbf{(NASGD)} and \textit{neural architecture search accelerated gradient descent} \textbf{(NASAGD)}. These algorithms are based on 1) the mathematical framework for semi-discrete optimization (deeply rooted in the geometric structure of optimal transport) that two of the authors have introduced and motivated in their theoretical work \cite{N1}, and 2) the neural architecture search methods originally proposed in \cite{Hillclimbing,Graphsearch}. We have chosen the \textit{network morphism} framework from\cite{Hillclimbing} because it allows us to illustrate the impact that our mathematical ideas can have on existing NAS algorithms without having to introduce the amount of background that other frameworks like those based on reinforcement learning would require.



The paper \cite{Hillclimbing} considers an iterative process where in a first step the parameters/weights of a collection of architectures are optimized for a \textit{fixed time}, and in a second step the set of architectures are updated by applying network morphisms to the best performing networks in the previous stage; these two steps are repeated until some stopping criterion is reached. These network morphisms are used to introduce local graphs of architectures.
While in our approaches we also use the concept of network morphism, the time spent in training a given set of networks is \textit{dynamically chosen} as determined by an evolving particle system. In our numerical experiments we observe that our algorithms change architectures much earlier than the fixed amount of time proposed in \cite{Hillclimbing}, while achieving error rates of $4.06\%$ for the CIFAR-10 data set trained over 12 hours with a single GPU for NASGD, and of $3.96\%$ on the same training data set trained over 1 day with one GPU for NASAGD. Given the shorter time spent exploring architectures, both of our methodologies can be set to consider positive and negative architecture morphisms/mutations, i.e. add/remove components to/from neural networks, as opposed to restricting to positive mutations as most approaches in the literature do.  We believe that the improved exploration of the architecture space inherent to our algorithms can potentially be exploited further by combining it with reinforcement learning techniques, but as we mentioned earlier this is out of the scope of this work. 
We emphasize again that while here we have restricted the use of the semi-discrete gradient dynamics introduced in \cite{N1} to the morphism framework from \cite{Hillclimbing}, we believe that our ideas have the potential to increase the speed at which architectures are explored in other NAS paradigms such as parameter sharing \cite{ParameterSharing} and differential architecture search \cite{DAS}. We will explore this direction in future work.

\noindent 

\nc

\subsection{Review of neural architecture search frameworks and related works}
\label{sec:literature}

There is an enormous literature on neural architecture search methodologies and some of its applications  (see
\cite{search_lit} for an overview on the subject). Roughly speaking, most methodologies found in the literature fall into the following families:

\begin{itemize}
	
	\item [-] \textbf{Reinforcement learning approaches.}
	The first group builds on ideas from reinforcement learning \cite{ReinfLearning} and uses the concept of a controller/agent, who designs architectures (takes actions), trains them, and then receives rewards based on the performance of the trained networks. As in a typical reinforcement learning problem, the controller aims at maximizing a discounted future reward function. In this setting, the search for optimal policies corresponds to the design of architectures through gradient-based optimization of the future reward function using model-free methodologies such as REINFORCEMENT \cite{REINFORCE}.  This reinforcement learning setting relies on two key components. Firstly, a neural network encoding/representation. This representation corresponds to deciding on a parameterization for the space of actions that the controller can take. Secondly, since the full training of a network is computationally expensive, different strategies need to be introduced to make the repeated evaluation of the reward function feasible. In the reinforcement learning setting from \cite{ReinfLearning} these strategies are fundamental given that the computation of a single policy gradient step may involve the training of several neural architectures.\\
	
	\item [-] \textbf{Evolutionary algorithmns.} The second major group of neural search methodologies is based on evolutionary algorithms \cite{RegularizedEvolution,NeuroEvolution}. As in the reinforcement learning setting, evolutionary algorithms rely on suitable architecture encodings that facilitate the specification of rules for merging and mutation of different neural architectures. In the evolutionary setting, issues with competing conventions for the merging of two parent networks may arise. In order to address these issues, papers like \cite{NeuroEvolution} propose genetic encodings that depend on historical markings of mutations as they occur in an evolving population of architectures. The NEAT methodology from \cite{NeuroEvolution} offers a solution to the problem of competing conventions in diverse topologies. In evolutionary algorithms, the issue of evaluating an expensive objective is also present. Indeed, the evaluation of the fitness function used to determine what ``individuals" should persist in time requires the full training of architecture models. Several strategies have been designed to address this issue. For example, a common technique is to evaluate the fitness function only at architecture candidates that are expected to return high fitness values. In turn, these candidates can be determined following strategies such as the covariance matrix evolution method from \cite{CMAEvolutionStartegy}. Originally this idea was applied in the context of neural networks for continuous hyperparameter optimization.
\end{itemize}

Some strategies used in the literature to lighten the computational burden of training multiple neural networks in NAS include: parameter sharing approaches  \cite{ParameterSharing}; methods used on specific application domains where architecture spaces are more concrete and thus specific strategies can be implemented as in \cite{ProgressiveNeural}; techniques focused on transferring knowledge from simpler learning problems to more difficult ones as in \cite{TransferableArchitectures}; strategies that use Bayesian model optimization, i.e. \textit{hyperparameter optimization} techniques such as SMBO as in \cite{ProgressiveNeural}; minimization of surrogate objectives as in \cite{Hyperparameter1,Hyperparameter2}.

\subsection{Outline}  We organize the rest of the paper as follows. In section \ref{ideal}, we introduce our algorithms NASGD and NASAGD. To motivate them, we first discuss two particle dynamics used for the optimization of a function defined on a semi-discrete space $\R^d \times \G$ for $\G= (\G, K)$ a finite weighted graph.  In section \ref{num_results} we compare the performance of NASGD and NASAGD against other NAS algorithms when working with the CIFAR-10 data set. In section \ref{fine_tuning} we provide more details on the implementation of our algorithms. In section \ref{conclusion} we provide some closing remarks and discuss future directions for research. 




\section{Our algorithms}\label{ideal}

In this section we introduce our algorithms NASGD and NASAGD. In order to motivate them, for pedagogical purposes we first consider an idealized setting where we imagine that NAS can be seen as a tensorized semi-discrete optimization problem of the form:
\begin{equation} \label{eq:semidiscreteopti}
\minn_{(x,g)\in\R^{d}\times\mathcal{G}}V(x,g).
\end{equation}
In the above, it will be useful to think of the $g$ coordinate as an \textit{architecture} and the $x$ coordinate as the \textit{parameters} of that architecture. It will also be useful to  think of $\G=(\G, K)$ as a finite similarity graph of architectures with $K$ a matrix of positive weights characterizing a neighborhood of a given architecture (later on $\G$ is defined in terms of network morphisms around a given architecture --see section \ref{morphisms}), and $V$ as a loss function (for concreteness cross-entropy loss ) which quantifies how well an architecture with given parameters performs in classifying a given training data set. Working in this ideal setting, in the next two subsections we introduce particle systems that aim at solving \eqref{eq:semidiscreteopti}. These particle systems are inspired by the gradient flow equations derived in \cite{N1} that we now discuss.


\subsection{First order algorithm}\label{first_order} The starting point of our discussion is a modification of equation (2.13) in \cite{N1} now reading:
\begin{align}\begin{aligned}\label{eqn:GradFlowEMod}\partial_{t}&f_t(x,g)=\text{div}_{x}(f_{t}(x,g)\nabla_{x}V(x,g))\\
+ & \sum_{g^{\prime}\in\mathcal{G}}\big[\log f_{t}(g)+V(x,g)-(\log f_t(g^{\prime})+V(x,g^{\prime}))\big]K(g,g^{\prime})\theta_{x,g,g'}(f_{t}(x,g),f_{t}(x,g')), \quad t>0.
\end{aligned}
\end{align}
In the above, $f_t(x,g)$ must be interpreted as a probability distribution on $\R^d \times \G$ and $f_t(g)$ as the corresponding marginal distribution on $g$. $\nabla_x$ denotes the gradient in $\R^d$ and $\text{div}_x$ the divergence operator acting on vector fields on $\R^d$.   
The second term on the right hand side of \eqref{eqn:GradFlowEMod} is a divergence term on the graph acting on graph vector fields which are nothing but real valued functions defined on the set of edges of the graph. The term $\theta_{x,g,g'}(f_t(x,g), f_t(x,g'))$ plays the role of interpolation between the masses located at the points $(x,g)$ and $(x,g')$, and it provides a simple way to define induced masses on the edges of the graph. With induced masses on the set of edges one can in turn define fluxes along the graph that are in close correspondence with the ones found in the dynamic formulation of optimal transport in the Euclidean space setting (see \cite{Maas}). 

The relevance of the evolution of distributions \eqref{eqn:GradFlowEMod} is that it can be interpreted as a continuous time \textit{steepest descent equation} for the minimization of the energy:  
\begin{equation}
\label{eqn:EntropyMarginal}
\widetilde{\mathcal{E}}(f):= \sum_{g \in \G}  \log f(g) f(g) + \sum_{g} \int_{\R^d} V(x,g) f(x,g)
\end{equation}
with respect to the geometric structure on the space of probability measures on $\R^d \times \G$ that was discussed in section 2.3 in \cite{N1}. Naturally, the choice of different interpolators $\theta$ endow the space of measures with a different geometry. In \cite{N1} the emphasis was given to choices of $\theta$ that give rise to a Riemannian structure on the space of measures, but alternative choices of $\theta$, like the one made in \cite{Finsler}, induce a general \textit{Finslerian} structure instead. In this paper we work with an interpolator inducing a Finslerian structure and in particular define
\begin{equation}
\theta_{x,g,g'}(s,s'):= s \mathds{1}_{U(x,g,g')>0} + s'\mathds{1}_{U(x,g,g') <0}, \quad s,s'>0,
\label{eqn:interpolator}
\end{equation}
where $U(x,g,g'):= \log f_{t}(g)+V(x,g)-(\log f_t(g^{\prime})+V(x,g^{\prime})) $.

\begin{remark}
	With the entropic term used in \eqref{eqn:EntropyMarginal} we only allow ``wandering" in the $g$ coordinate. This term encourages exploration of the architecture space.
\end{remark}

We now consider a collection of moving particles on $\R^d \times \G$ whose evolving empirical distribution aims at mimicking the evolution described in \eqref{eqn:GradFlowEMod}. Initially the particles have locations $(x_i, g_i)$ $i=1, \dots, N$ where we assume that if $g_i=g_j$ then $x_i=x_j$ (see Remark \ref{rem:Initial} below). For fixed time step $\tau>0$, particle locations are updated by repeatedly applying the following steps:

\begin{itemize}

	\item[-] \noindent \textbf{Step 1: Updating parameters (Training)}: For each particle $i$ with position $(x_i,g_i)$ we update its parameters by setting:
	\[
	x_i^{\tau}=x_i-\tau\nabla_{x}V(x_i,g_i).
	\]
\end{itemize}

\begin{itemize}
	\item [-] \noindent \textbf{Step 2: Moving in the architecture space (Mutation)}: First, for each of the particles $i$ with position $(x_i,g_i)$ we decide to change its $g$ coordinate with probability:
	\[\tau\sum_{j}(\log f(g_j)+V(x_j,g_j))-(\log f(g_i)+V(x_i,g_i))^{-}K(g_i,g_j) ,	\]
	or $1$ if the above number is greater than $1$. If we decide to move particle $i$, we move it to the position of particle $j$, i.e. $(x_j,g_j)$, with probability $p_j$:
	\begin{align*}\begin{aligned}\hspace{5em} p_j \propto \big[ & \log f(g_j)+V(x_j,g_j)-(\log f(g_i)+V(x_i,g_i))\big]^{-}K(g_i,g_j).
	\end{aligned}
	\end{align*}
	In the above, $f(g)$ denotes the ratio of particles that are located at $g.$ Additionally, $a^{-}= \max \{0, -a\}  $ denotes the negative part of the quantity $a$.

\end{itemize}


\begin{remark}\label{rem:Initial}
	Given the assumptions on the initial locations of the particles, throughout all the iterations of Step 1 and Step 2 it is true that if $g_i=g_j$ then $x_i=x_j$.  This is convenient from a computational perspective because in this way the number of architectures that need to get trained is equal to the number of nodes in the graph (which is small) and not to the number of particles in our scheme. 
\end{remark}






\begin{remark}
	\label{rem:PorousMedium}
	By modifying the energy $\widetilde{\mathcal{E}}(f)$ replacing the entropic term with an energy of the form $\frac{1}{\beta+1}\sum_g (f(g))^{\beta+1} $ for some parameter $\beta>0$, one can motivate a new particle system where in Step 2 every appearance of $\log f$ is replaced with $f^\beta$. The effect of this change is that the resulting particle system moves at a slower rate than the version of the particle system as described in Step 2. 
\end{remark}

\subsection{Second order algorithm}
\label{secondorder:sec}
Our \textit{second order algorithm} is inspired by the system of equations (2.17) in \cite{N1} which now reads:
\begin{align}\label{second_order}\begin{cases}
\partial_t f_t(x,g)+\text{div}_x( f_t(x,g)\nabla_x\varphi_t )+ \sum_{g'}(\varphi_t(x,g')-\varphi_t(x,g) )K(g,g')\theta_{x,g,g'}(f_t(x,g), f_t(x,g))=0\\
\partial_{t}\varphi_t+\frac{1}{2}|\nabla_x\varphi_t|^{2}+\sum_{g^{\prime}}\big(\varphi_t(x,g)- \varphi_t(x,g')\big)^{2}K(g,g')\partial_{s}\theta_{x,g,g'}(f_t(x,g),f_t(x,g'))\\
\hspace{20em}= -[ \gamma \varphi_t(x,g) +\log f_t(g)+V(x,g)], \quad t>0.
\end{cases}\end{align}
We use $\theta$ as in \eqref{eqn:interpolator} except that now we set $U(x,g,g'):= \varphi_t(x,g')-\varphi_t(x,g)$. System \eqref{second_order} describes a second order algorithm for the optimization of $\widetilde\EE$ -- see sections 2.4 and 3.3. in \cite{N1} for a detailed discussion. Here, the function $\varphi_t$ is a real valued function over $\R^d \times \G$ that can be interpreted as \textit{momentum} variable. $\gamma\geq 0$ is a friction parameter.

System \eqref{second_order} motivates the following particle system, where now we think that the position of a particle is characterized by the tuple $(x_i,g_i,v_i)$ where $x_i,v_i \in \R^d$, $g_i \in \G$, and in addition we have a potential function $\varphi : \G \rightarrow \R$ that also gets updated. Initially, we assume that if $g_i=g_j$ then $x_i=x_j$ and $v_i=v_j$. We also assume that initially $\varphi$ is identically equal to zero.


We summarize the \textit{second order gradient flow dynamics} as the iterative application of three steps:
\begin{itemize}
	\item[-]\noindent \textbf{Step 1: Updating parameters (Training):} For each particle $i$ located at  $(x_i,g_i,v_i)$ we update its parameters $x_i,v_i$ by setting
	\begin{align*}\begin{aligned}x^{\tau}_i & =x_i+\tau v_i,\\
	v_i^{\tau} & =v_i-\tau(\gamma v_i+\nabla_{x}V(x_i,g_i)).\\
	\end{aligned}
	\end{align*}
\end{itemize}

\begin{itemize}
	\item[-] \textbf{Step 2: Moving in the architecture space (Mutation):} First, for each of the particles $i$ with position $(x_i,g_i,v_i)$ we decide to  move it with probability 
	\[\tau\sum_{j}(\varphi(g_i)-\varphi(g_j))^{-} K(g_i,g_j),\]
	or $1$ if the above quantity is greater than $1$. Then, if we decided to move the particle $i$ we move it to location of particle $j$, $(x_j,g_j,v_j)$ with probability $p_j$
	\begin{align*}\begin{aligned} p_j\propto  (\varphi(g_i)-\varphi(g_j))^{-} K(g_i,g_j).  
	\end{aligned}
	\end{align*}
\end{itemize}

\begin{itemize}
	\item [-] \textbf{Step 3: Updating  momentum on the $g$ coordinate}:
	We update $\varphi$ according to:
	
	\begin{align*}\begin{aligned}
	\varphi^{\tau}(g_i)=\varphi(g_i)-\frac{\tau}{2}|v_i|^{2}&-\tau\bigg(\sum_{j}\big(\big[\varphi(g_i)-\varphi(g_j)\big]^{-}\big)^{2}K(g_i,g_j)\bigg) -\tau(\gamma \varphi(g_i)+\log f(g_i)+V(x_i,g_i)),\\
	\end{aligned}\end{align*}
	for every particle $i$. Here, $f(g)$ represents the ratio of particles located at $g$.  
\end{itemize}
\begin{remark}
	Notice that given the assumption on the initial locations of the particles, throughout all the iterations of Step 1 and Step 2 and Step 3 we make sure that if $g_i=g_j$ then $x_i=x_j$ and $v_i=v_j$.\end{remark}



\subsection{NASGD and NASAGD}
\label{morphisms}

We are now ready to describe our algorithms {NASGD} and {NASAGD}:

\begin{algorithm}
	\caption{NASGD}
	\label{alg:algorithm-label}
	\begin{enumerate}
		\item[1.] Load an initial architecture $g_0$ with initial parameters $x_0$ and set $r=0$.
		\item[2.] Construct a graph $\mathcal{G}_r$ around $g_r$ using the notion of network morphism introduced in \cite{Hillclimbing}. More precisely, we produce $n_{neigh}$ new architectures with associated parameters, each new architecture is constructed by modifying $g_r$ using a \textit{single} network morphism from \cite{Hillclimbing}. Then define $\mathcal{G}_r$ as the set consisting of the loaded $n_{neigh}$ architectures and the architecture $g_r$. Set the graph weights $K(g,g')$ (for example, setting all weights to one). 
		
		\item [3.] Put $N$ particles on $( x_r,g_r)$ and put $1$ ``ghost" particle on each of the remaining architectures in $\G_r$. The architectures for these ghost particles are never updated (to make sure we always have at least one particle in each of the architectures in $\G_r$), but certainly their parameters will. 
		
		Then, run the dynamics discussed in section \ref{first_order} on the graph $\G_r$ (or the modified dynamics see, Remark \ref{rem:PorousMedium} and Appendix \ref{ap:heuristic} of the ArXiv version of this paper \cite{ArxivUS}) until the node in $\G_r \setminus \{ g_r \}$ with the most particles $g^{max}$ has twice as many particles as $g_{r}$.

		Set $r=r+1$. Set $g_{r}= g^{max}$ and $x_{r}= x^{max}$, where $x^{max}$ are the parameters of architecture $g^{max}$ at the moment of stopping the particle dynamics. 
		
		\item[4.] If size of $g_r$ exceeds a prespecified threshold (in terms of number of convolutional layers for example) go to 5. If not go back to 2.
		
		\item[5.] Train $g_r$ until convergence.
		
	\end{enumerate}

\end{algorithm}

\begin{algorithm}[H]
	\caption{NASAGD}
	\label{alg:algorithm-label}
	\begin{enumerate}
		\item[1.] Load an initial architecture $g_0$ with initial parameters $x_0$ and $v_0$. Set $\varphi_0\equiv 0$. Set $r=0$.
		\item[2.] Construct a graph $\mathcal{G}_r$ around $g_r$ using the notion of network morphism introduced in \cite{Hillclimbing}. More precisely, we produce $n_{neigh}$ new architectures with associated parameters, each new architecture is constructed by modifying $g_r$ using a \textit{single} network morphism from \cite{Hillclimbing}. Then define $\mathcal{G}_r$ as the set consisting of the loaded $n_{neigh}$ architectures and the architecture $g_r$. Set the graph weights $K(g,g')$ (for example, setting all weights to one).
		
		\item [3.] For NASGD: Locate $N$ particles on $( x_r,g_r,v_r),$ and put $1$ ``ghost" particle on each of the remaining architectures in $\G_r$. The architectures for these ghost particles are never updated (to make sure we always have at least one particle in each of the architectures in $\G_r$), but certainly their parameters will. 
		
		Initialize the potential $\varphi_r$ to 0. Then run the dynamics discussed in section \ref{second_order} (or the modified dynamics see, Remark \ref{rem:PorousMedium} and Appendix \ref{ap:heuristic} of the ArXiv version of this paper \cite{ArxivUS}) until the node in $\G_r \setminus \{ g_r \}$ with the most particles $g^{max}$ has twice as many particles as $g_{r}$. 
		
		Set $r=r+1$. Set $g_{r}= g^{max}$, $x_{r}= x^{max}$, $v_r= v^{max}$ where $x^{max}, v^{max}$ are the parameters of architecture $g^{max}$ at the moment of stopping the particle dynamics. 
		
		\item[4.] If size of $g_r$ exceeds a prespecified threshold (in terms of number of convolutional layers for example) go to 5. If not go back  to 2.
		\item [5.] Train $g_r$ until convergence.
		
	\end{enumerate}

\end{algorithm}

\section{Experiments}\label{num_results}

\noindent In this section, we present our numerical results. Before displaying them, we compare in more detail our algorithms with the ones originally proposed in \cite{Hillclimbing} and \cite{Graphsearch}. We begin by summarizing their framework and explaining how we apply our gradient flow dynamics to it. 


\begin{itemize}
	\item[-]\textbf{Hill climbing and graph search framework:} The algorithm proposed in \cite{Hillclimbing} starts with a small pre-trained network. The authors then suggest an application of a fixed number $n_{NM}$ of random network morphisms to this base network to produce $n_{neigh}$ children networks. They train the children networks during $epochs_{neigh}$ epochs using SGD and the cosine annealing strategy introduced in \cite{WR}. The learning rate is interpolated from $\lambda_{start}$ to $\lambda_{final}$ during the $epochs_{neigh}$ epochs. Their algorithm iterates this process, restarting it $n_{steps}$ times. At the beginning of each cycle, it applies the same number of morphisms to the best performing architecture in the previous cycle to produce new $n_{neigh}$ children architectures. After the $n_{steps}$ cycles, the best network is trained until convergence using the same range for the cosine aliasing. The work \cite{Graphsearch} builds upon the framework introduced in \cite{Hillclimbing} by including linear morphisms and gradient weighting techniques that prevent old layers from overfitting .
	
\end{itemize}

\noindent Next, we describe how we use our fist and second-order architecture search algorithms NASGD and NASAGD to perform our numerical experiments.

\begin{itemize}
	
	\item[-] \textbf{Our framework for NASGD and NASAGD}:   In a similar way to \cite{Hillclimbing} and \cite{Graphsearch}, we pre-train an initial network $g_0$ with the structure Conv-MaxPool-Conv-MaxPool-Conv-Softmax for 20 epochs using cosine aliasing that interpolates between 0.5 and $10^{-7}$. We use $g_{0}$ with parameters $x_{0}$ as the initial data for our gradient flow dynamics introduced in section \ref{first_order} for the first-order algorithm NASGD and section \ref{second_order} for the second-order algorithm NASAGD.   During the NASGD and NASAGD algorithms, we use cosine aliasing interpolating the learning rate from $\lambda_{start}$ to $\lambda_{final}$  with a restart period of $epochs_{neigh}.$ In contrast to the NASH approach from \cite{Hillclimbing}, since we initialize new architectures, we do not reset the time step along with the interpolation for the $epochs_{neighs}$ epochs. Our particle system \textit{dynamically} determines the number of initialization.  We continue this overall dynamics, resetting the learning rate from $\lambda_{start}$ to $\lambda_{final}$ every $epochs_{neigh}$  at most $n_{steps}$ times. We perform several experiments letting the first and second-order gradient flow dynamics run for different lengths of time. Finally, we train the found architectures until convergence.  
\end{itemize}  
\begin{remark} 
	Here, by Conv, and throughout the rest of the paper we mean:
	\[\text{Conv=Conv+batchnorm+Relu}.\]
\end{remark}

\subsection{Models found by our NASGD and NASAGD algorithms.} In our numerical experiments, we find two models NASGD1 and NASAGD1 (see Appendix \ref{App:NASGD} and \ref{App:NASAGD} in the ArXiv version of this paper \cite{ArxivUS}). We found the model NASGD1 running the first-order gradient flow dynamics with $n_{steps}=0.89.$  On the other hand, we obtain the model  NASAGD1, by running the second-order gradient flow dynamics with $n_{steps}=2.54.$  In the table below, we display the rest of the parameters used to find these models. We also present the corresponding parameters used to find the NASH2 model obtained from \cite{Hillclimbing} and the NASGraph model from \cite{Graphsearch}:

\begin{center}
	\begin{tabular}{ |c|c|c|c|c| } 
		\hline
		\textbf{Variable} & \textbf{NASH2} & \textbf{NASGraph} & \textbf{NASGD1}  & \textbf{NASAGD1} \\ 
		$n_{steps}$ & 8 & 10 & 0.89 & 2.54\\ 
		$n_{NM}$ & 5 & 5 & \textit{dynamic} & \textit{dynamic}\\
		$n_{neigh}$ & 8 & 8 & 8 & 8\\
		$epoch_{neigh}$ & 17 & 16 & 18 & 18\\
		$\lambda_{start}$ & 0.05 & 0.1 & 0.05 & 0.05\\
		$\lambda_{final}$ & 0 & 0 & $10^{-7}$ & $10^{-7}$\\
		Gradient Stopping & No & Yes & No & No\\
		\hline
	\end{tabular}
\end{center}

Here, $n_{steps}$ denotes the number of restart cycles for the cosine aliasing; $n_{NM}$ is the number of morphism operations applied on a given restart cycle; $n_{neigh}$ is the number of children architectures generated every time the current best model changes; $epoch_{neigh}$ is the number of epochs that go by before the cosine aliasing is restarted; $\lambda_{final}$ and $\lambda_{start}$ are the parameters required for SGDR.

\subsection{Numerical Results}

We performed two experiments to produce NASGD1  and NASAGD1, which respectively used the first and second order algorithms to learn from the CIFAR-10 data set. We compare the performance of our methodologies against NASH2 and NasGraph in the next table:


\begin{center}
	\begin{tabular}{ |p{2.5cm}||p{2.5cm}|p{2.5cm}|p{2.5cm}|p{2.5cm}|  }
		\hline
		\multicolumn{5}{|c|}{Numerical Experiments} \\
		\hline
		CIFAR 10 & \textbf{model} &  \textbf{resources} & \textbf{\# params $\times 10^6$} & \textbf{error}\\
		\hline
		
		&   NASH2  & 1GPU, 1day   & 19.7 & 5.2\\
		&NASGraph & 1GPU, 20 h& ? & 4.96\\
		& NASGD1 & 1GPU, 12h& 25.4 & 4.06\\
		&   NASAGD1  & 1GPU, 1 day&  22.9 & 3.96\\
		\hline
	\end{tabular}
\end{center}

Besides producing better accuracy rates, it is worth highlighting that our algorithms can explore many more architectures (about a 40 times more) than in \cite{Hillclimbing} with the same computational resources. We took advantage of this faster exploration and considered positive as well as negative architecture mutations, i.e., mutations that can increase or decrease the number of filters, layers, skip, and dimension of convolutional kernels.

\section{Fine-tuning of the first and second-order gradient flow dynamics}\label{fine_tuning}

\noindent In this section we explain the main points that need to be modified from the gradient flow dynamics discussed in sections \ref{first_order} and \ref{secondorder:sec} in order to make them better suited for the NAS problem. In particular, these adaptations are implemented in order to make our algorithms viable for learning from the CIFAR-10 data set. For the reader's convenience, we have stated the equations for the final version of our dynamics in Appendix \ref{ap:heuristic} of the ArXiv version of this paper \cite{ArxivUS}. We fine-tune these dynamics by addressing the following points:

\subsection{Stochastic gradient descent and decoupling of the loss function:}
The first adaptation we make is to consider a stochastic gradient descent version of the dynamics proposed in sections \ref{first_order} and \ref{secondorder:sec}. To perform stochastic gradient descent we first need a sequence of mini-batches $\{X_{k}\}_{k}$ (of the same size) that are randomly chosen from our labeled training data set. We denote the batch size with $S_{X}$. In our experiments, we set $S_{X}=64.$ In each  iteration of our dynamics, for the training step we use one of these mini-batches $X_{k}$ (we technically use different mini-batches for each architecture, but for notational simplicity, let us pretend it is the same for all architectures in each iteration) in order to define the loss function to be optimized (more details below). We use a different sequence of mini-batches $\{Y_{k}\}_{k},$ which we build from a set that is \textit{disjoint} from the one used to build the $\{ X_k \}_{k}$. These mini-batches are used in the validation of the network performance. In our implementation, we set the size of these validation mini-batches to be $S_{Y}=32$. The data in the $Y_k$ are never used to update network parameters.


More precisely, we do the following:
\begin{itemize}
	\item[(a)] \textit{Adaptation of the training step: } During the training step of our algorithm, when we are updating the parameters of an architecture $g$ with parameters $x$, we denote the value of the forward feed of $g$ evaluated at the mini-batch $X_{k}$ with $V_{k}(x,g).$ Then, at the $k$-th step we update the parameters of the network $(x,g)$ with the gradient $\nabla_{x}V_{k}(x,g)$. 
	\item[(b)] \textit{Adaptation of the mutation step: } For this step we use the validation mini-batches  $\{Y_{k}\}_{k}.$ That is, given an architecture $g$ with parameters $x$ that is loaded on the $k$-th iteration, we denote the value of the forward feed of $g$ evaluated at the mini-batch $Y_{k}$ by $\tilde{V}_{k}(x,g).$ In Appendix \ref{ap:heuristic} of the ArXiv version of this paper \cite{ArxivUS} we rewrite our mutation steps in terms of $\tilde{V}$ valuations. We notice that in the hill-climbing framework \cite{Hillclimbing}, the authors use the accuracy on the validation set to determine the best model. Similarly, we define $\tilde{V}$ using mini-batches sampled from a validation set (we reiterate that we do not use these mini-batches in the training step). In this way, the mutation step assigns particles to those architectures that are better at generalizing their training to sets that are disjoint from the training set. As expected, we found that this increases test accuracy and prevents overfitting.
\end{itemize}

\subsection{Warm restarts and final adaptation of the dynamics:} As discussed in \cite{Hillclimbing}, the training step discussed in section \ref{first_order} is generally not fast enough to produce significant variations between architectures in a reasonable time.  Hence, for the experiment NASGD1, we have used the technique in \cite{WR} and perform the training with momentum and cosine aliasing. In Appendix \ref{ap:heuristic} from \cite{ArxivUS} we denote the size of the time step determined by the cosine aliasing in the $k$-th iteration by $\tau_{k}$. 

For the mutation steps in both the first and second order models we followed Remark \ref{rem:PorousMedium} in order to slow down the propagation of particles. We found that this choice produced better results than the original particle system described in sections \ref{first_order} and \ref{secondorder:sec} -- see Appendix \ref{ap:heuristic} in \cite{ArxivUS} for a more detailed formulation of our implemented particle system. Also, for the mutation step of the second-order model, we do not use the term $\frac{1}{2}|v_{i}|^{2}$ since it is quite expensive to compute and empirically was not observed to affect our results significantly.

\subsection{Constraining the graph of architectures:} We adopt a similar strategy to \cite{Graphsearch} to dynamically modify the size of the last layer of our network before the fully connected part. Additionally, we impose global constraints on loaded architectures such as a maximum number of pool layers and filters. We also constrain the number of incoming connections that a given layer may receive.

\begin{remark}In contrast to \cite{Hillclimbing}, we apply a single morphism at the same time. We can afford this because our particle dynamics sample more architectures overall. On the other hand, this choice was somewhat arbitrary, and it could be productive to randomize the number of morphisms applied in order to produce new competing architectures at the beginning of each cycle. Additionally, we have considered morphisms that may reduce the size of our network. Architectures produced by negative morphisms are occasionally preferred by the particle dynamics.
\end{remark}


\section{Conclusions and discussion}\label{conclusion}

In this work we have proposed novel first and second order gradient descent algorithms for neural architecture search (NASGD and NASAGD). The theoretical gradient flow structures in the space of probability measures over a semi-discrete space introduced in \cite{N1} serve as the primary motivation for our algorithms. Our numerical experiments show the practical viability of our methodologies: we achieve competitive results while analyzing a considerably larger amount of architectures using the same computational resources as other methodologies.

The methodologies introduced in this paper are part of a first step in a broader program where we envision the use of well defined mathematical structures to motivate new learning algorithms that use neural networks. Although here we have achieved competitive results, we believe that there are still several possible directions for improvement that are worth exploring in the future. 

First, we believe that a more careful analysis of hyperparameters in our methods is needed. The number of morphisms, the structure of local graphs (i.e. the choice of $K$), and the mutation coefficients, were all fixed in a reasonable way and were not calibrated neither theoretically nor using a data driven method.



Secondly, given that our methods have the added advantage that they can analyze a much larger number of architectures with the same resources used by the approaches in \cite{Hillclimbing} and \cite{Graphsearch}, we believe that our methods have the potential to develop a strong synergy with reinforcement learning approaches. In reinforcement learning methods, a controller learns from data collected as architectures are explored. As data gets collected faster, a trained controller may in principle better predict morphism chains, improving in this way the construction of local graphs. In general, we believe that the analytical and geometric ideas that motivated the concrete methodologies presented in this work have the potential to increase the speed at which architectures are explored in other NAS paradigms. Some examples of such paradigms include parameter sharing \cite{ParameterSharing} and differential architecture search \cite{DAS}.

A final research direction that stems from our work is motivated by the task of neural network \textit{distillation}. Indeed, by considering negative morphisms only, we may actually use the same particle dynamics used to explore local graphs in order to prune complex neural networks in search of simpler ones with good accuracy rates. This direction will be explored in future work.

\bibliography{ref}
\bibliographystyle{abbrv}	
\appendix
\label{App:NASGD}
\section{Architechture found by the  NASGD algorithm}

\begin{center}
	\includegraphics[scale=0.65]{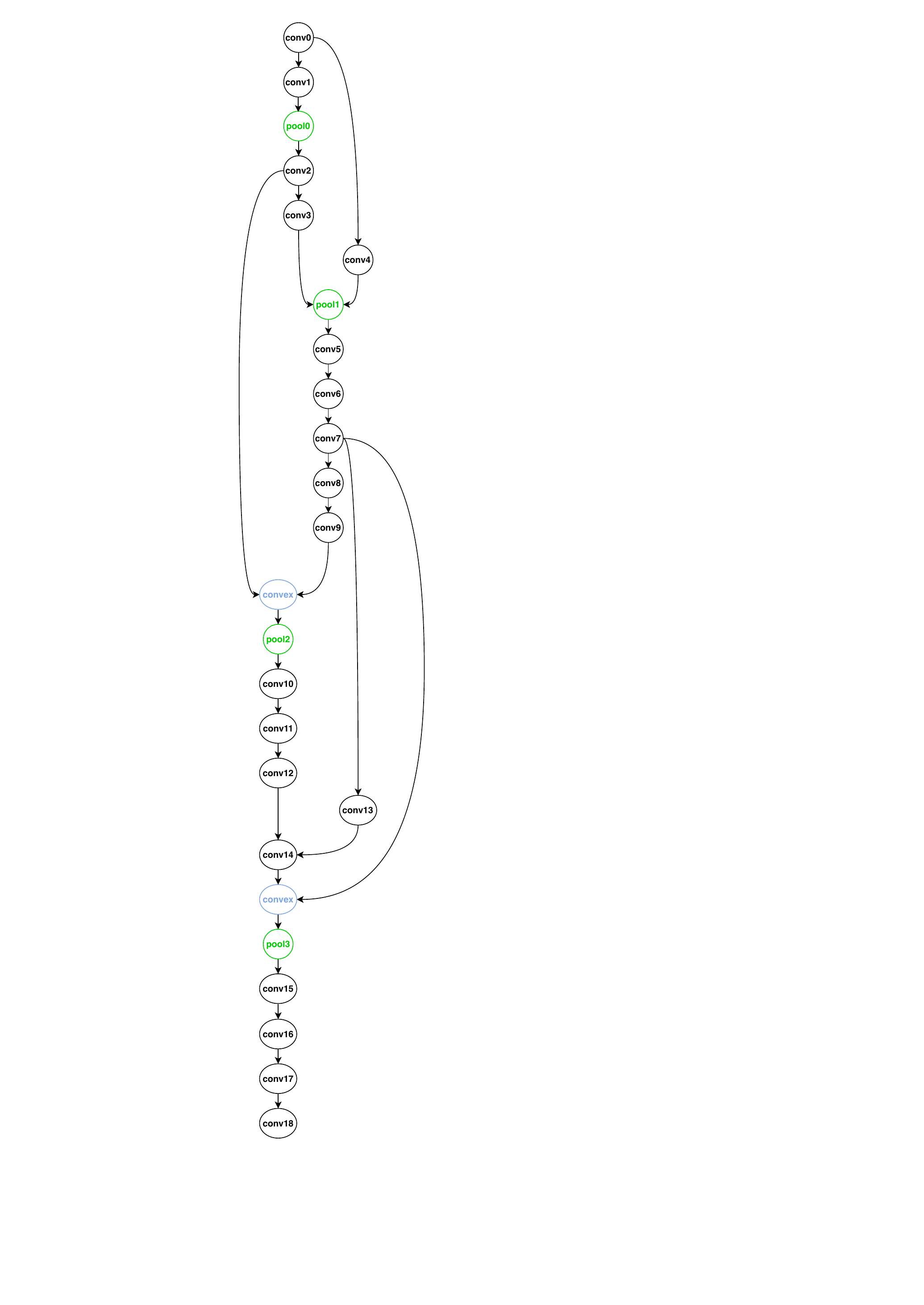}
\end{center}

\section{Architechture found by the  NASAGD algorithm}
\label{App:NASAGD}

\begin{center}
	\includegraphics[scale=0.65]{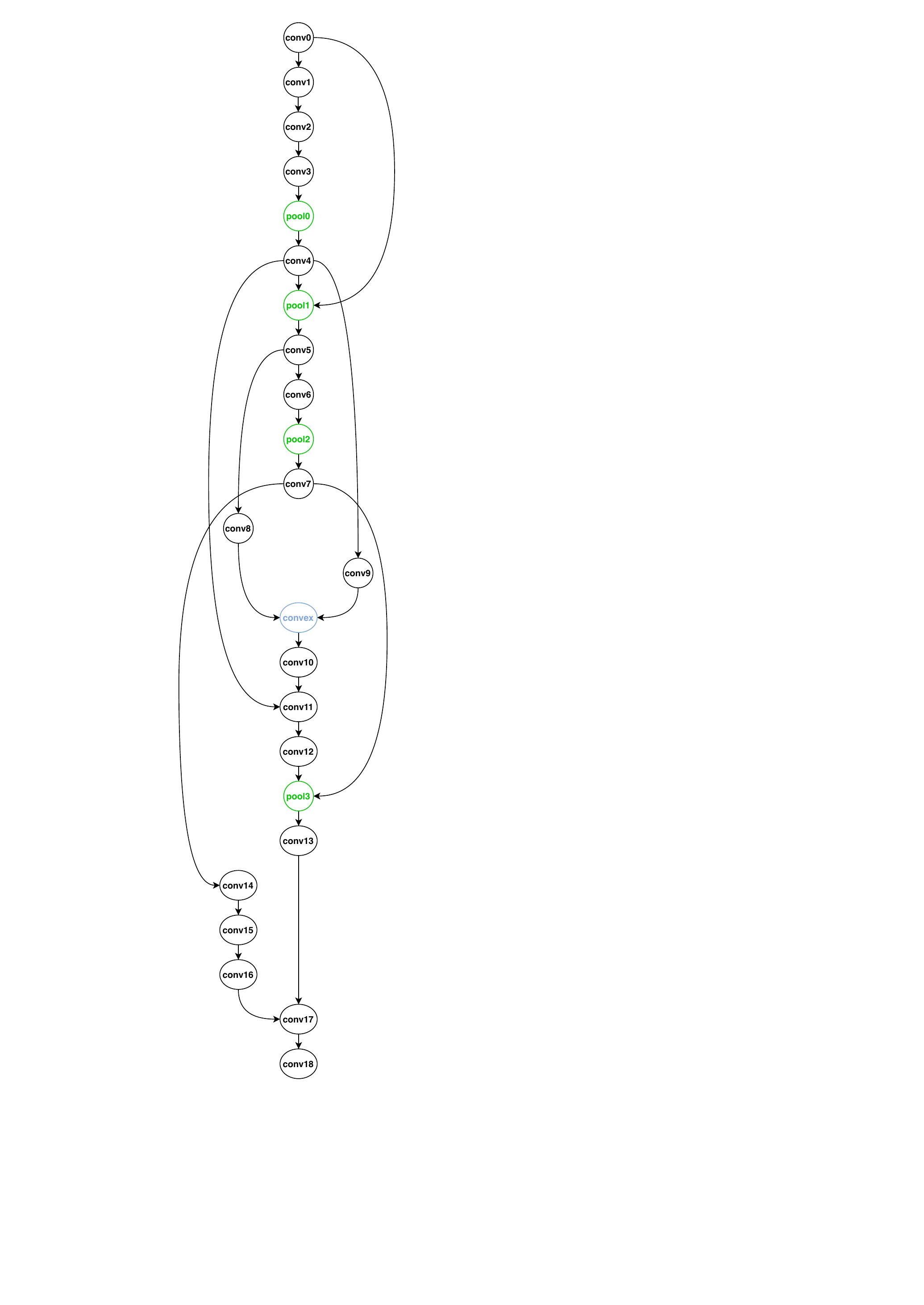}
\end{center}

\section{Learning curve for the NASGD1 and NASAGD1 experiments}

\begin{center}
	\begin{figure}[H]
		\includegraphics[scale=0.5]{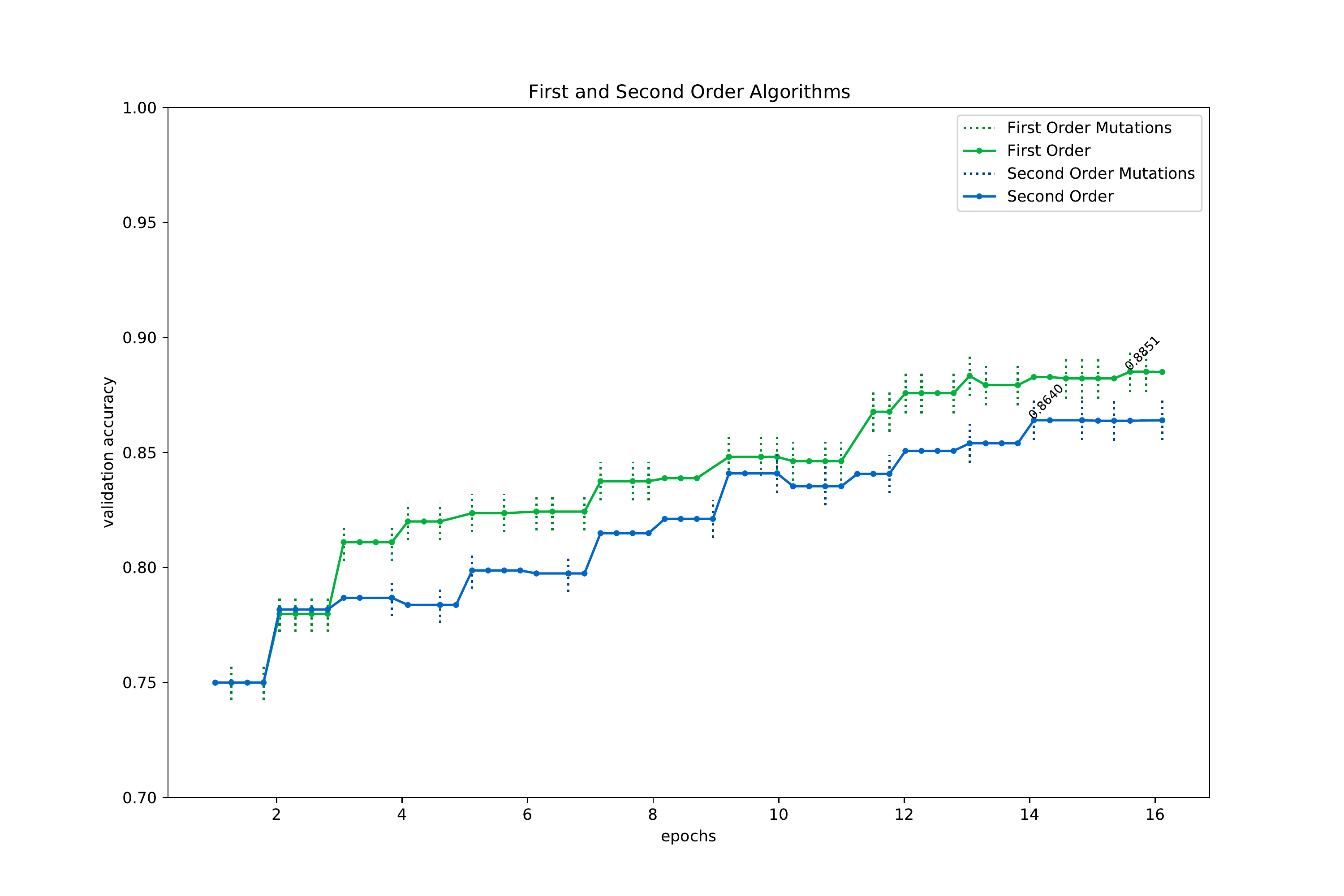}
		\caption{On this plot, we display the accuracy of the current best-performing epochs during  $n_{steps}=0.89$ cycles of the architecture search part of the algorithm. NASGD finds our model during this time and then trains it until it reaches the error rate of 4.06.}
	\end{figure}
\end{center}

\section{Implemented particle methods}\label{ap:heuristic}

\noindent In this section, for the convenience of the reader, we write the final equations for the adapted dynamics from sections \ref{first_order} and \ref{secondorder:sec} after all the points discussed in section \ref{fine_tuning} are considered.

\subsection{First order algorithm (\textbf{NASGD})} We summarize the first order gradient flow dynamics as the iterative application of the following two steps:\\

\begin{itemize}

	\item[-] \noindent \textbf{Step 1: Updating the parameters (Training)}: For each of the previously initialized networks $(x_{i},g_{i},v_{i})$ we compute  $V_{k}(x_{i},g_i),$ and $\nabla_x V_k(\cdot,g_i),$ and update its parameters by setting
	\begin{align*}\begin{aligned}x_{i}^{\tau_{k}} & =x_{i}+\tau_{k}v_{i},\\
	v_{i}^{\tau_{k}} & =v_{i}-\tau_{k}(v_{i}+\nabla_x V_k (x_i,g_{i})).
	\end{aligned}
	\end{align*}
	Here, and henceforth, we use $\tau_{k}$ to denote the length of the time step determined by the global cosine aliasing interpolation. Similarly, $V_k(g_{i},x))$ denotes the valuation function of the current mini-batch for the stochastic gradient descent. Note that as described in section \ref{fine_tuning} we use second order dynamics for the training even though technically speaking we are stating our first order approach.\\
\end{itemize}

\begin{itemize}
	\item [-] \noindent \textbf{Step 2: Moving the particles (Mutation)}: First, for each particle at each initialized network $(x,g)$, we select to  move it, with probability
	\[\kappa\tau_{k}\sum_{g'}[(f^{\beta}(g^{\prime})+\tilde{V}_{k}(x^{\prime},g^{\prime}))-(f^{\beta}(g)+\tilde{V_{k}}(x,g))]^{+}K(g,g'),	\]
	or $1$ if the above quantity is greater than $1$. Here, $\kappa>0$ denotes the mutation coefficient. As discussed in section \ref{fine_tuning},  $\tilde{V}_{k}$ is used to denote the running average of the valuation of the lost function in the respective mini-batches of a given architecture.\\

	\noindent Then, if we decide to move  a particle in $(x,g)$ we move it to $(x^{\prime},g^{\prime})$ with probability, 
	\begin{align*}\begin{aligned}\hspace{5em}\bigg(\big[ & f^{\beta}(g^{\prime})+\tilde{V}_{k}(x^{\prime},g')-(f^{\beta}(g)+\tilde{V}_{k}(x,g))\big]^{+}K(g,g')\bigg)\\
	& \quad/\bigg(\sum_{g''}\big[f^{\beta}(g^{\prime\prime})+\tilde{V}_{k}(x^{\prime\prime},g'')-[f^{\beta}(g)+\tilde{V}_{k}(x,g)]\big]^{+} K(g,g'')\bigg).
	\end{aligned}
	\end{align*}

\end{itemize}

\subsection{Second order algorithm (\textbf{NASAGD}):} \label{second_detailed} We summarize the second order gradient flow dynamics as the iterative application of the following three steps:\\\\

\begin{itemize}
	
	\item[-]\noindent \textbf{Step 1: Updating the parameters (Training):} For each of the previously initialized networks $(x_{i},g_{i},v_{i})$ we compute  $V_{k}(x_{i},g_{i})$, compute its gradient $\nabla_x V_k(\cdot,g_{i}),$ and update its parameters by setting
	\begin{align*}\begin{aligned}x_{i}^{\tau_{k}} & =x_{i}+\tau_{k}v_{i},\\
	v_{i}^{\tau_{k}} & =v_{i}-\tau_{k}(v_{i}+\nabla_{x}V_{k}(x_{i},g_{i})).
	\end{aligned}
	\end{align*}

\end{itemize}

\begin{itemize}
	\item[-] \textbf{Step 2: Moving the particles (Mutation)}:  First, for each particle at $(x,g,v)$, we move it with probability
	\[\kappa\hspace{1mm}\tau_{k}\sum_{g'}(\varphi(g)-\varphi(g^{\prime}))^{-}K(g,g')	,\]
	or $1$ if the above quantity is greater than $1$.
	\noindent Then, if we decided to move the particle, we move it to $(x^{\prime},g^{\prime},v')$ with probability :
	\begin{align*}\begin{aligned}(\varphi(g) - \varphi(g^{\prime}))^{-}K(g,g^{\prime})\times\bigg(\sum_{g''}\big(\varphi(g)-\varphi(g^{\prime\prime})\big)^{-} K(g,g^{\prime\prime})\bigg)^{-1}.\end{aligned}
	\end{align*}
\end{itemize}

\begin{itemize}
	\item [-] \textbf{Step 3: Updating the velocity field (applying external force)}:
	To do this, we set
	\begin{align*}\begin{aligned}\varphi^{\tau_{k}}(g)=\varphi(g) & -\tau_{k}\bigg(\sum_{K(g^{\prime},g)}\big(\big[\varphi(g)-\varphi(g^{\prime})\big]^{-}K(g,g^{\prime})\big)^{2}\bigg)\\
	& \hspace{3em}-\tau_{k}(f^{\beta}(g)+\tilde{V}_{k}(x,g)),
	\end{aligned}
	\end{align*}
	for each previously initialized architecture $g.$\\
	
	\noindent During our implementation, we found it useful to restart $\varphi$ and set it equal to $0$ every time that the quantity
	\[ \sum_{g,g^{\prime}}(\varphi(g)-\varphi(g^{\prime}))^{-}\cdot(\tilde{V}_{k}(x,g)-\tilde{V}_{k}(x^{\prime},g^{\prime}))^{-}K(g,g^{\prime})f(g),
	\]
	becomes positive. Heuristically, this quantity measures the rate of change of the average loss function for all the particles as they evolve in time.
\end{itemize}

\end{document}